\documentclass[10pt,twocolumn,letterpaper]{article}

\usepackage[pagenumbers]{cvpr} 

\usepackage{graphicx}
\usepackage{amsmath}
\usepackage{amssymb}
\usepackage{booktabs}
\usepackage{multirow}

\usepackage[pagebackref,breaklinks,colorlinks]{hyperref}

\usepackage[capitalize]{cleveref}
\crefname{section}{Sec.}{Secs.}
\Crefname{section}{Section}{Sections}
\Crefname{table}{Table}{Tables}
\crefname{table}{Tab.}{Tabs.}

\def\cvprPaperID{0016} 
\def\confName{CVPR}
\def\confYear{2023}

\begin{document}

\title{EvenNICER-SLAM: Event-based Neural Implicit Encoding SLAM}


\author{Shi Chen$^{1}$ \quad\quad Danda Pani Paudel$^{1,2}$ \quad\quad Luc Van Gool$^{1,2}$\\
$^1$Computer Vision Lab, ETH Zurich \quad\quad $^2$INSAIT, Sofia University
}
\maketitle

\begin{abstract}
The advancement of dense visual simultaneous localization and mapping (SLAM) has been greatly facilitated by the emergence of neural implicit representations. Neural implicit encoding SLAM, a typical example of which is NICE-SLAM, has recently demonstrated promising results in large-scale indoor scenes. However, these methods typically rely on temporally dense RGB-D image streams as input in order to function properly. When the input source does not support high frame rates or the camera movement is too fast, these methods often experience crashes or significant degradation in tracking and mapping accuracy. In this paper, we propose EvenNICER-SLAM, a novel approach that addresses this issue through the incorporation of event cameras. Event cameras are bio-inspired cameras that respond to intensity changes instead of absolute brightness. Specifically, we integrated an event loss backpropagation stream into the NICE-SLAM pipeline to enhance camera tracking with insufficient RGB-D input. We found through quantitative evaluation that EvenNICER-SLAM, with an inclusion of higher-frequency event image input, significantly outperforms NICE-SLAM with reduced RGB-D input frequency. Our results suggest the potential for event cameras to improve the robustness of dense SLAM systems against fast camera motion in real-world scenarios.
\end{abstract}

\begin{figure}[t]
\centering
\includegraphics[scale=0.33]{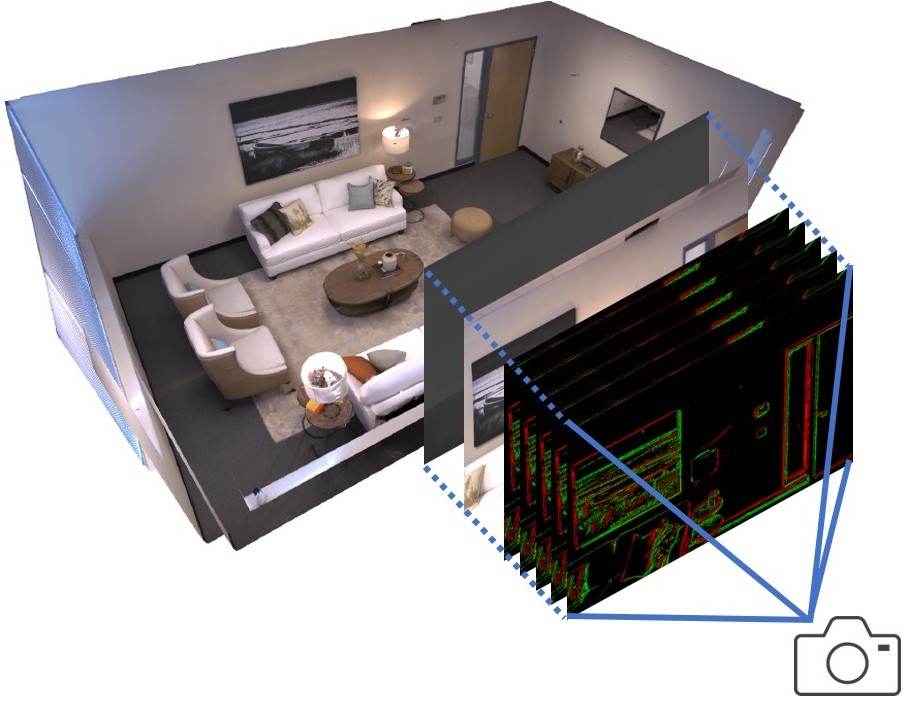}
\caption[Cover Image]{A brief illustration of EvenNICER-SLAM. Our method takes advantage of the high-temporal-resolution feature of event data to facilitate camera tracking in high-speed application scenarios. EvenNICER-SLAM typically takes lower-frequency RGB-D input and higher-frequency event image input and significantly outperforms its predecessor, NICE-SLAM\cite{zhu2022nice}, in both camera tracking and mapping.}
  \label{fig:cover}
\end{figure}

\section{Introduction}
\label{sec:intro}

Dense visual Simultaneous Localization and Mapping (SLAM) has been a popular subfield of 3D computer vision for its industrial applications in autonomous vehicle navigation, mixed/augmented reality (MR/AR), etc. While many dense SLAM studies are based on traditional methods \cite{klein2009parallel, newcombe2011dtam, schops2019bad}, there has been a shift towards learning-based approaches \cite{bloesch2018codeslam, zhi2019scenecode, sucar2020nodeslam} in recent years. Particularly, the incorporation of Neural Radiance Field (NeRF) \cite{mildenhall2021nerf} has been transformative for its capability to represent scenes with decent photometric accuracy. IMap \cite{sucar2021imap} was the first real-time dense SLAM method that encodes scenes into neural implicit representations. NICE-SLAM \cite{zhu2022nice}, built on the advancement of iMap, further enabled accurate camera tracking and mapping in larger indoor scenes by optimizing multi-level spatially partitioned feature grids (See Sec. \ref{niceslam}).

Despite its success with spatially dense camera poses in indoor scenes, NICE-SLAM \cite{zhu2022nice} is prone to failure in cases of large inter-frame camera translation or rotation, which occurs when the camera moves quickly in relation to the frame rate. We found through experiments that while NICE-SLAM works well with a full sequence from the Replica dataset \cite{straub2019replica}, it mostly crashes when the same sequence is only fed every fifth frame. To be more specific, we can do a rough estimation of avarage camera pose translation between adjacent frames. According to our observations, the camera travels a total distance of around 20 meters throughout a Replica sequence of 2000 frames. That converts to roughly 1cm/frame, and 5cm/frame with the reduced input. In other words, NICE-SLAM would likely crash with a Replica RGB-D sequence where the camera travels an average of 5cm every frame.

In scenarios involving high-speed camera motion, however, the requirements are significantly higher. Autonomous racing drones, for example, are known to fly at maximum speeds of over 30m/s in recent years \cite{hanover2023autonomous}. As most RGB-D cameras operate at 30 frames per second (FPS), the maximum camera pose translation between two consecutive frames would be around 1 meter. This is two orders of magnitude larger than the estimated inter-frame translation for the failure cases of NICE-SLAM \cite{zhu2022nice}. Additionally, there are also scenarios where cameras regularly experience high-speed rotation, such as mixed reality headsets. The Microsoft Hololens 2, for example, is a representative device of this type of application. According to the technical report of Hololens 2 \cite{ungureanu2020hololens}, the equipped depth camera can only operate at a maximum of 5 FPS for high resolution ($1024\times1024$, similar to the $1200\times680$ resolution in Replica \cite{straub2019replica}) depth map output. Nevertheless, Guan et al. \cite{guan2022deepmix} pointed out that the peak speed of human head movement can reach 240 degrees per second, which means the maximum rotation per frame would be about 48 degrees in this case. In contrast, we do not observe such large inter-frame rotations even in the reduced Replica sequence. Based on the inference above, it is clear that NICE-SLAM, as one of the recent works in the field of neural implicit encoding SLAM, is not currently applicable in high-speed scenarios.

In an effort to address the above limitation of NICE-SLAM \cite{zhu2022nice}, we considered the recent incorporation of event cameras \cite{gallego2020event} in computer vision research. Event cameras have high temporal resolution, high dynamic range, and low latency, making them particularly suitable for high-speed applications. Further information regarding event cameras can be found in Sec. \ref{sec:event}.

In this paper, we propose EvenNICER-SLAM, a novel dense SLAM system that utilizes lower-frequency RGB-D input and higher-frequency event image input. EvenNICER-SLAM is built upon the NICE-SLAM \cite{zhu2022nice} framework by adding a new event loss backpropagation stream. We conduct quantitative evaluations in camera tracking and mapping, and our results demonstrate that EvenNICER-SLAM has a large advantage in both sections over NICE-SLAM with the same reduced RGB-D streams as input. Note that we reduce the frame frequency to virtually enlarge the inter-frame camera pose translation and rotation. To the best of our knowledge, we are the first to show that the integration of event input can boost neural implicit encoding SLAM with low RGB-D frame rates in both camera tracking and scene mapping. We believe that EvenNICER-SLAM will unlock the potential of neural implicit encoding SLAM in high-speed applications.

\section{Related Work}
\label{sec:related}

In this section, we first present multiple previous works on dense visual SLAM and NeRF (Neural Radiance Field) in the context of SLAM. Then, we proceed to offer an introduction of event cameras and event-based SLAM.

\subsection{Dense Visual SLAM}
Recent studies in the field of visual SLAM have primarily adopted the decoupled parallel structure of tracking and mapping,  which was first introduced in the seminal work PTAM \cite{klein2009parallel}. Building upon the sparse approach of PTAM, DTAM \cite{newcombe2011dtam} was developed to achieve dense visual SLAM in a similar decomposed manner. 

An alternative approach to directly optimizing depth maps has been proposed by several studies \cite{bloesch2018codeslam, zhi2019scenecode, sucar2020nodeslam}, in which the optimization is performed on a latent scene representation that can be decoded into dense geometry. This approach alleviates the previously expensive computational cost and has subsequently led to a boost in the practicality of dense visual SLAM, particularly in terms of real-time capability. Furthermore, it can also be seen as a prototype of recent works on NeRF-driven SLAM, which we will present later in Sec.~\ref{nerfslam}.

\begin{figure*}[t]
  \centering
  \includegraphics[width=\linewidth]{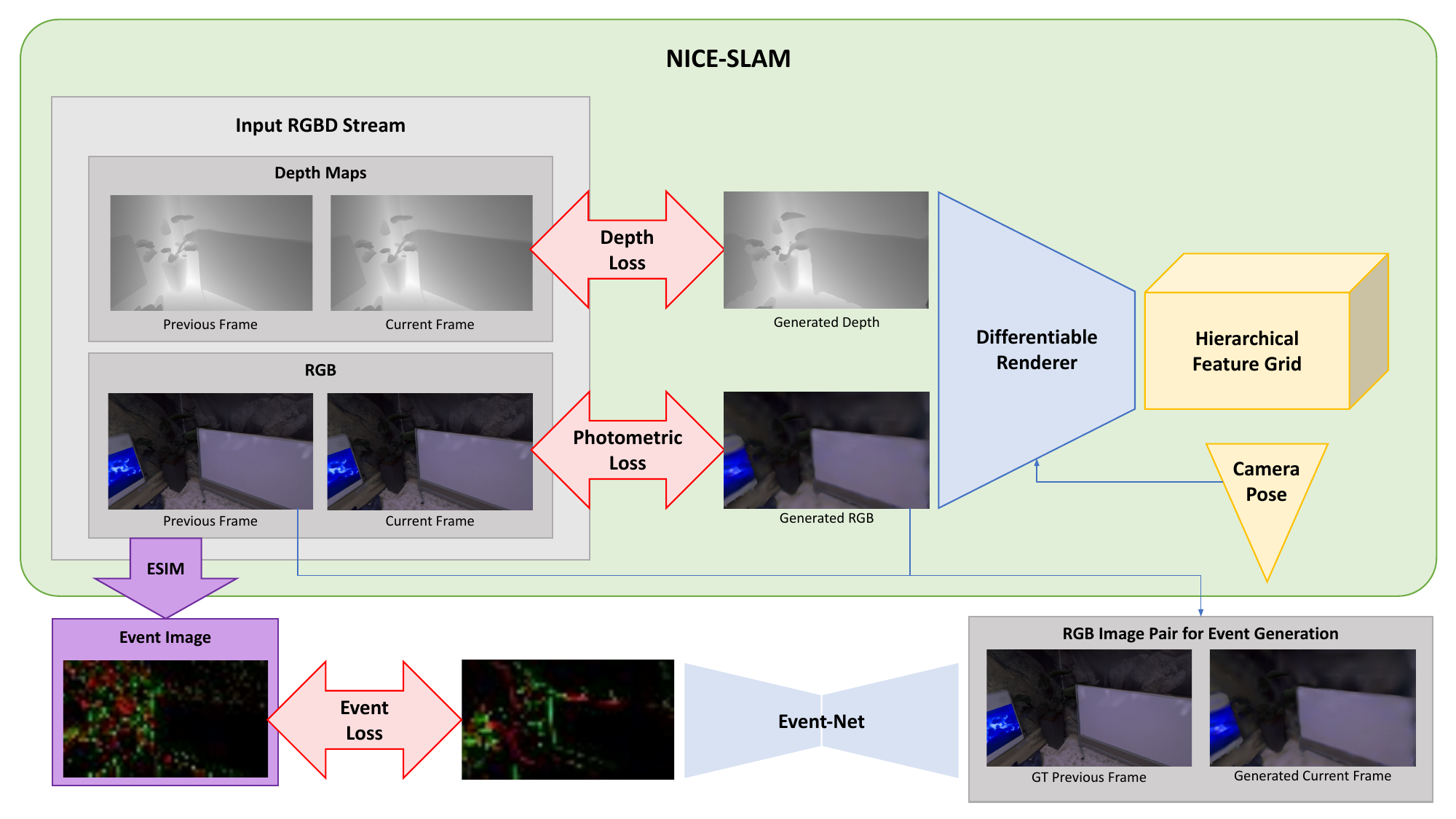}
  \caption[Overview of EvenNICER-SLAM]{An overview of EvenNICER-SLAM framework. Before running EvenNICER-SLAM, we first generate ground truth event images between all adjacent frames in the original dataset with ESIM \cite{rebecq2018esim}. During a run, we feed Event-Net with the currently rendered RGB image and the latest available ground truth RGB image prior to the current timestamp, in order to predict an event image between them. Accordingly, we sum up ground truth event images at corresponding timestamps and compute an event loss. Finally, the event loss is backpropagated through the differentiable Event-Net and renderer to optimize camera poses. The components within the green box are kept the same as in NICE-SLAM \cite{zhu2022nice}.
  }
  \label{fig:1}
\end{figure*}

\subsection{NeRF in the context of SLAM}
\label{nerfslam}

Among methods of implicit scene representation, NeRF \cite{mildenhall2021nerf} (Neural Radiance Field) has been the most extensively used one in recent years. It utilizes a large Multi-Layer Perceptron (MLP) that takes a spatial location and a viewing direction as input and outputs the corresponding occupancy and radiance to implicitly represent a scene. 

The view-dependent nature of NeRF has triggered preceding efforts to regress camera poses with an existing NeRF representation \cite{yen2021inerf}, which achieves a crucial prerequisite for dense visual SLAM. Additionally, a method for simultaneously fitting camera poses and a NeRF model was proposed in \cite{lin2021barf}, bringing NeRF even closer to practical application in SLAM. However, the optimization step of a single large MLP remained overly slow for online SLAM.

IMap \cite{sucar2021imap} was the first to achieve real-time dense visual SLAM using RGB-D input. However, iMap still utilizes a single MLP to represent the entire scene, which limits its performance in large scenes. NICE-SLAM \cite{zhu2022nice}, which advances the developments of iMap, addresses this limitation by encoding the scene with a hierarchical feature grid and predicting occupancy and colors with multi-level decoders. Our work is based on NICE-SLAM, and more details of its mechanism are introduced in Sec. \ref{niceslam}. Recently, NeRF-SLAM \cite{rosinol2022nerf} has shown  real-time dense SLAM with only RGB image sequences as input.

Despite these advances, none of the forementioned dense SLAM methods are robust against fast moving cameras and resulting motion blur. In this paper, we aim to tackle this problem by integrating event input into existing NeRF-based SLAM systems.

\subsection{Event Cameras and Event-based SLAM}
\label{sec:event}
Event cameras are a type of bio-inspired camera in which each pixel responds to increases and decreases of observed logarithmic brightness asynchronously with low latency \cite{gallego2020event}. The attractive properties of event cameras, such as high temporal resolution, high dynamic range, and reduced motion blur, have made them an ideal substitute for traditional RGB cameras in various computer vision tasks involving fast camera motion.

However, real event cameras are currently still expensive and thus cannot be used extensively. To address this issue, ESIM \cite{rebecq2018esim, Gehrig_2020_CVPR}, an open event camera simulator, was developed to make up for the insufficient event data. The ESIM is an essential component of our method. The usage of ESIM in our pipeline is introduced in Sec. \ref{sec:esim}.

Event-based SLAM is a relatively new topic that has been gaining popularity in recent years due to the benefits of event cameras mentioned above. Both Kim et al. \cite{kim2016real} and Rebecq et al. \cite{rebecq2016evo} proposed real-time SLAM with only events as input. However, both methods are only able to produce semi-dense 3D point clouds as the reconstruction of scenes. In contrast, our EvenNICER-SLAM makes use of neural implicit encoding to achieve dense reconstruction. More recently, Vidal et al. \cite{vidal2018ultimate} proposed a powerful SLAM system with hybrid input of events, standard RGB images, and inertial measurements. In comparison, our we exploit neural representation for dense mapping without requiring Inertial Measurement Unit (IMU) to track camera poses.

\section{Methods}

In this section, we first give an overview of the proposed EvenNICER-SLAM framework. Then, we delve into the specific methods implemented in each module of the system. Finally, we present our optimization method using event loss.

\subsection{Overview}
We provide an overview of the EvenNICER-SLAM framework in Fig. \ref{fig:1}. EvenNICER-SLAM is mainly based on NICE-SLAM \cite{zhu2022nice}, with the main difference being the reduction in the frame rate of RGB-D images fed to the SLAM, and the substitution of event images (as described in Section \ref{sec:esim}) at the original frame rate. 

ESIM \cite{rebecq2018esim, Gehrig_2020_CVPR}, an event camera simulator, is a substitute for real event cameras in the synthetic setting. We use it before running EvenNICER-SLAM to synthesize event images between all adjacent frames. Additionally, we introduce Event-Net, a differentiable event generator, to predict event images based on RGB images rendered by the SLAM. Finally, an event loss is calculated and backpropagated to optimize camera poses.

\subsection{ESIM and Event-Net}
\label{sec:esim}
Prior to conducting experiments on the SLAM system, we first synthesize event images between all adjacent frames from the original RGB image streams using ESIM \cite{rebecq2018esim, Gehrig_2020_CVPR}. ESIM is an open-source event camera simulator that takes two RGB images as input and simulates events in between. We integrate the simulated events in each time interval into an event image, which records the number of positive and negative events that are triggered at each pixel during a given period of time. We refer to these event images generated by ESIM as ``ground truth" event images. Note that we let ESIM operate in the log-irradiance domain. Specifically, irradiance $E$ (integer value ranging from 0 to 255) is converted to log-irradiance,
    \begin{equation}
        L = \ln{\frac{E}{255+\epsilon}},
        \label{eq:1}
    \end{equation}
where $\epsilon$ is a parameter to assure numerical stability.

Following the generation of event images, we pretrain Event-Net to take two consecutive RGB images as input and output an event image between them. The reason for adopting a neural network to learn the behavior of ESIM \cite{rebecq2018esim, Gehrig_2020_CVPR} is that our pipeline requires differentiability to backpropagate the event loss introduced in Sec. \ref{sec:optim}. In addition to utility ease, it potentially offers cross-domain  knowledge transfer by learning.
We employ the U-Net architecture \cite{ronneberger2015u} for Event-Net due to its capability to capture high resolution features. We specify the numbers of input and output channels as 6 (two images, three channels each) and 2 (one for positive events and the other for negative events), respectively.

\subsection{NICE-SLAM}
\label{niceslam}
NICE-SLAM \cite{zhu2022nice} takes an RGB-D stream as input and optimizes a scene representation and camera poses in an online fashion. The scene representation is implemented through a combination of hierarchical feature grids and multi-level pretrained decoders. Occupancy and color predictions from these decoders are then processed through a volume renderer to generate depth maps and RGB images. Subsequently, depth loss and photometric loss are computed and backpropagated to optimize the hierarchical feature grid and camera poses. In our work, we retain this setting as it is in NICE-SLAM.

The NICE-SLAM \cite{zhu2022nice} system is parallelized into three threads: coarse-level mapper, mid-\&fine-level mapper, and camera tracker. Among them, the camera tracker runs on every frame, while the mapper threads are dispatched at lower frequencies (typically every fifth frame). Currently, event loss backpropagation (Sec. \ref{sec:optim}) is only implemented in the camera tracker thread.

\begin{figure}[t]
\centering
\includegraphics[scale=0.19]{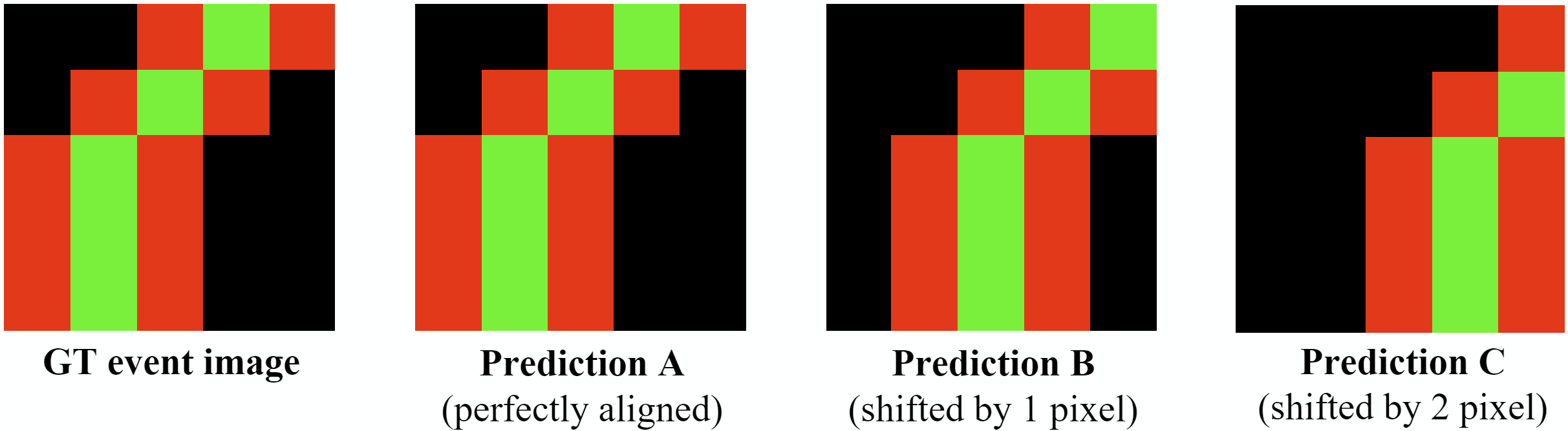}
\caption[Alignment Issue]{A simplified demonstration of event loss alignment issue. A red/green pixel represents a positive/negative event. In terms of prediction error, intuitively we would consider $A < B < C$. However, if we implement the event loss as a simple pixelwise error, the result will be a counterintuitive $A < C < B$.}
  \label{fig:alignment}
\end{figure}

\subsection{Optimization Using Event Loss}
\label{sec:optim}

With event loss, we aim to measure the discrepancy between predicted and ground truth event images. Ideally, Event-Net should generate an event image that is similar to the corresponding one from ESIM \cite{rebecq2018esim, Gehrig_2020_CVPR}, assuming that the SLAM predicted the transition of camera pose accurately. In practice, however, it is challenging to generate perfectly-aligned event images. Due to the high-frequency nature of event images, any pixelwise misalignment between predicted and ground truth event images can cause an irregular change of a naive pixelwise event loss (see Figure \ref{fig:alignment}). This can make the parameter space overly rough and undermine optimization stability. To smooth the parameter space and stabilize the optimization process, we apply a Gaussian filter to both predicted and ground truth event images before obtaining an $L_{2}$ loss between them.

Another challenge is that rendering a full-resolution image every iteration is time-consuming and requires considerable computational resources. To address this, we only render a largely downsampled image each iteration and downscale its corresponding event image accordingly.

\begin{figure}[t]
\centering
\includegraphics[scale=0.24]{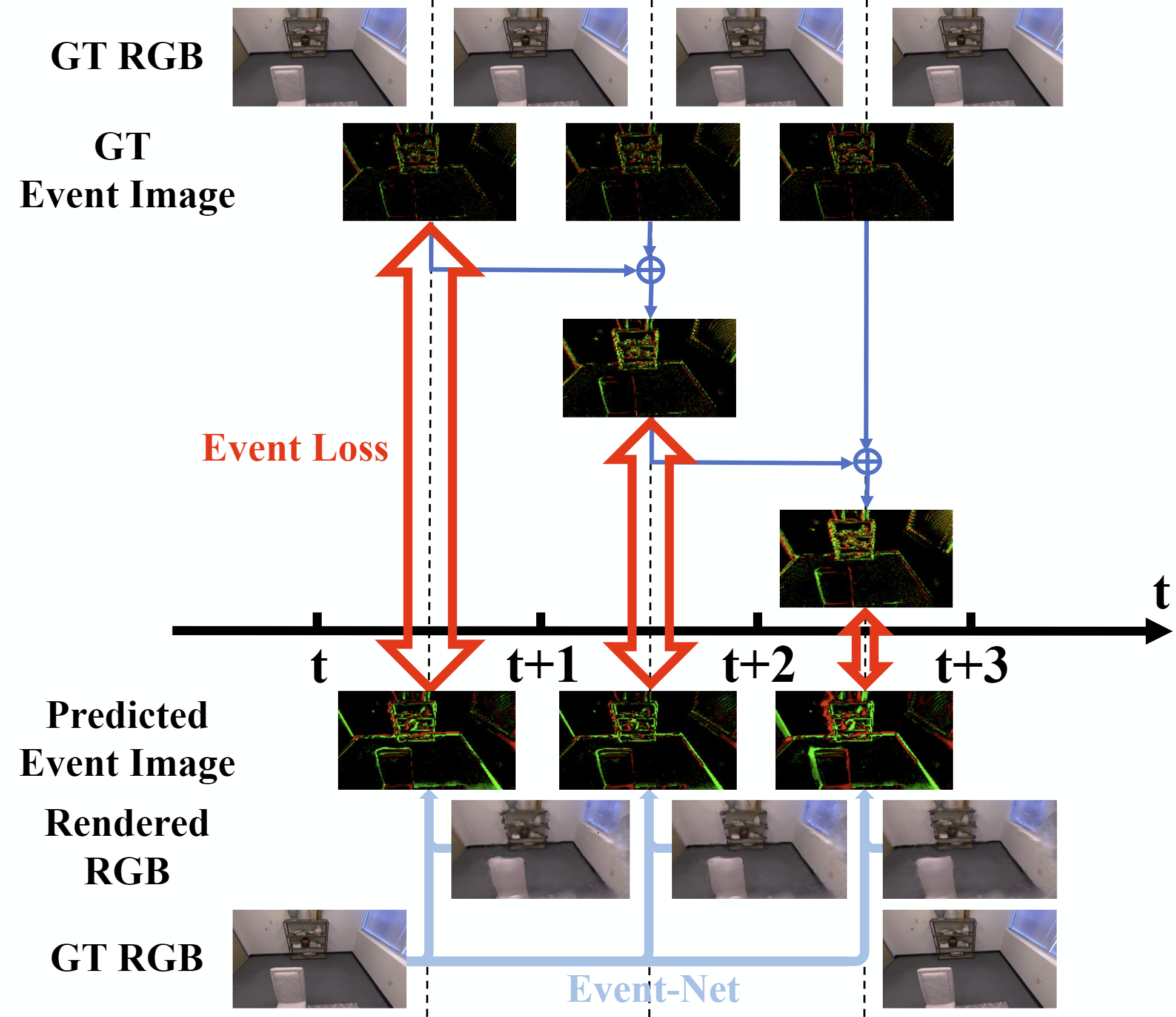}
\caption[timeline]{The timeline of one tracking cycle of EvenNICER-SLAM with frame gap $\tau = 3$. The preprocessing part is illustrated above the time axis, and below the axis are the actual processes executed by EvenNICER-SLAM during a run.}
  \label{fig:timeline}
\end{figure}

In Figure \ref{fig:timeline}, we illustrate the timeline of a tracking cycle of EvenNICER-SLAM. In the experimental setting of EvenNICER-SLAM, ground truth RGB-D images are only fed when the mapper threads run and prohibited otherwise. On the other hand, a ground truth event images is fed for every frame. To avoid error accumulation, we always use the latest available ground truth RGB image prior to the current timestamp instead of a rendered one as the first input of Event-Net. Note that it is downscaled to the same size as that of the rendered current RGB image before inference. Accordingly, we simply sum up the corresponding ground truth event images in between.

In summary, we have predicted event images:
\begin{equation}
    \widehat{I}_{event}^{t} = f(I_{RGB}^{t_{GT}}, \widehat{I}_{RGB}^{t}),\, t-\tau\leq t_{GT} \leq t-1,
    \label{eq:2}
\end{equation}
where $\widehat{I}_{event}$ is the predicted event image, $f$ stands for the mapping from input to output of Event-Net, $I_{RGB}$ and $\widehat{I}_{RGB}$ are (downscaled) ground truth and rendered RGB images respectively, $t$ is the current timestamp (frame index), $\tau$ is the ``frame gap", namely the period of ground truth RGB-D images being fed in relation to event images (by default $\tau = 5$), and $t_{GT}$ is the most recent timestamp when ground truth RGB-D images are fed. Finally, we calculate the $L_{2}$ event loss:
\begin{equation}
\begin{split}
    L_{event} = \lambda_{event}\sum_{x,y}\Bigg(Gaussian\bigg(\sum_{t=t_{GT}+1}^{t}{I}_{event}^{t}(x, y)\bigg)\\
    -Gaussian\bigg(\widehat{I}_{event}^{t}(x, y)\bigg)\Bigg)^2,
\end{split}
    \label{eq:3}
\end{equation}
where ${I}_{event}$ is the ground truth event image, and $\lambda_{event}$ is a coefficient balancing the event loss and reconstruction loss (the sum of depth loss and photometric loss).

Finally, the event loss is backpropagated through the differentiable Event-Net and renderer to optimize camera poses. We fix the parameters of Event-Net during the entire run. The learning rates are maintained the same as those of the reconstruction loss in NICE-SLAM \cite{zhu2022nice}.

\begin{figure*}[t]
\centering
  \begin{subfigure}[b]{0.49\textwidth}
    \centering
    \includegraphics[keepaspectratio, scale=0.60]{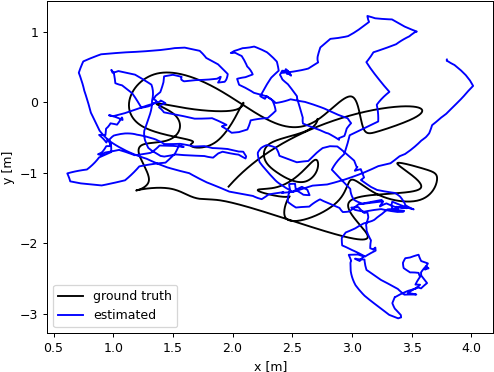}
    \caption{NICE-SLAM \cite{zhu2022nice}}
  \end{subfigure}
  \hfill
  \begin{subfigure}[b]{0.49\textwidth}
    \centering
    \includegraphics[keepaspectratio, scale=0.60]{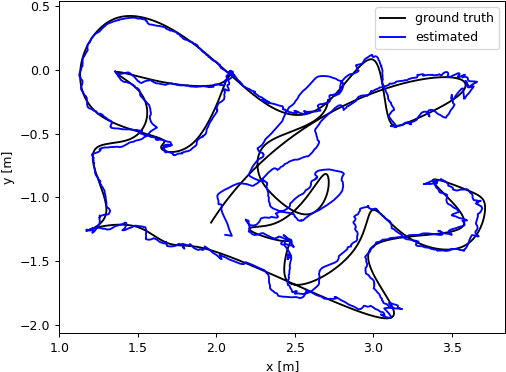}
    \caption{\textbf{EvenNICER-SLAM}}
  \end{subfigure}
   \caption[Examples: trajectories]{Comparison of camera trajectories estimated by NICE-SLAM\cite{zhu2022nice} and EvenNICER-SLAM (\texttt{room2} from Replica\cite{straub2019replica}). The frame gap of RGB-D images is set to $\tau = 5$. With the extra event supervision, EvenNICER-SLAM tracks the camera motion with a higher accuracy.}
  \label{fig:traj}
\end{figure*}

\begin{figure*}[t]

\centering
  \begin{subfigure}[b]{0.48\linewidth}
    \centering
    \includegraphics[keepaspectratio, scale=0.18]{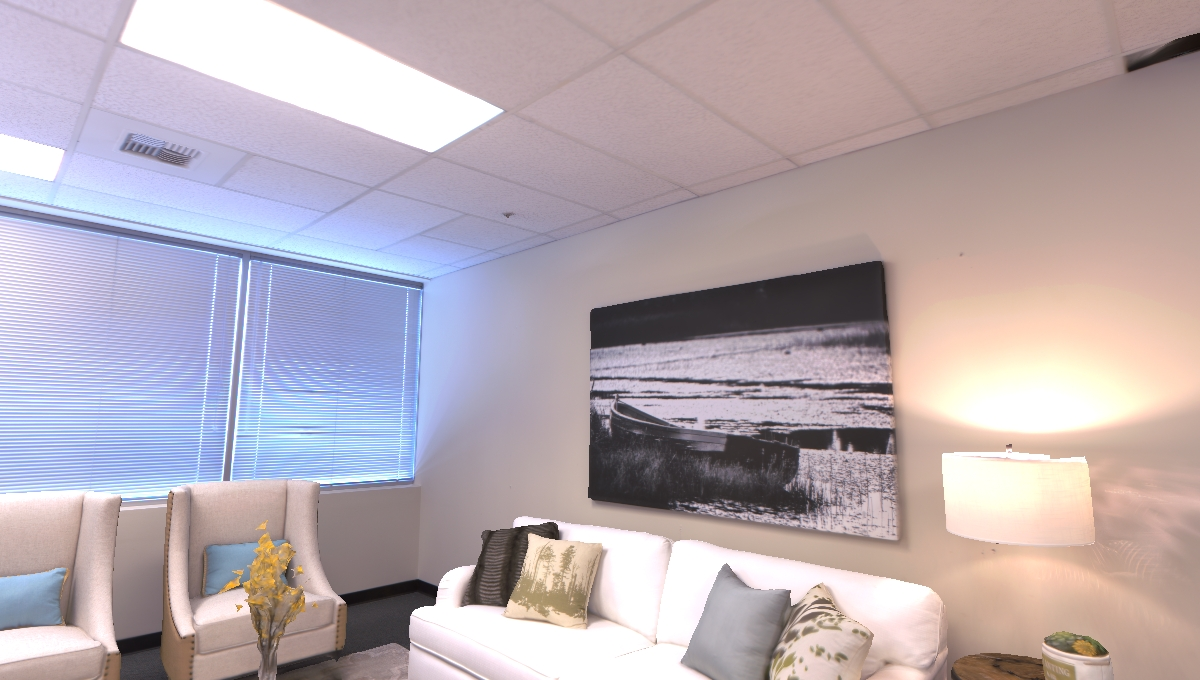}
    \caption{Ground truth}
  \end{subfigure}
  \hfill
  \begin{subfigure}[b]{0.48\linewidth}
    \centering
    \includegraphics[keepaspectratio, scale=0.18]{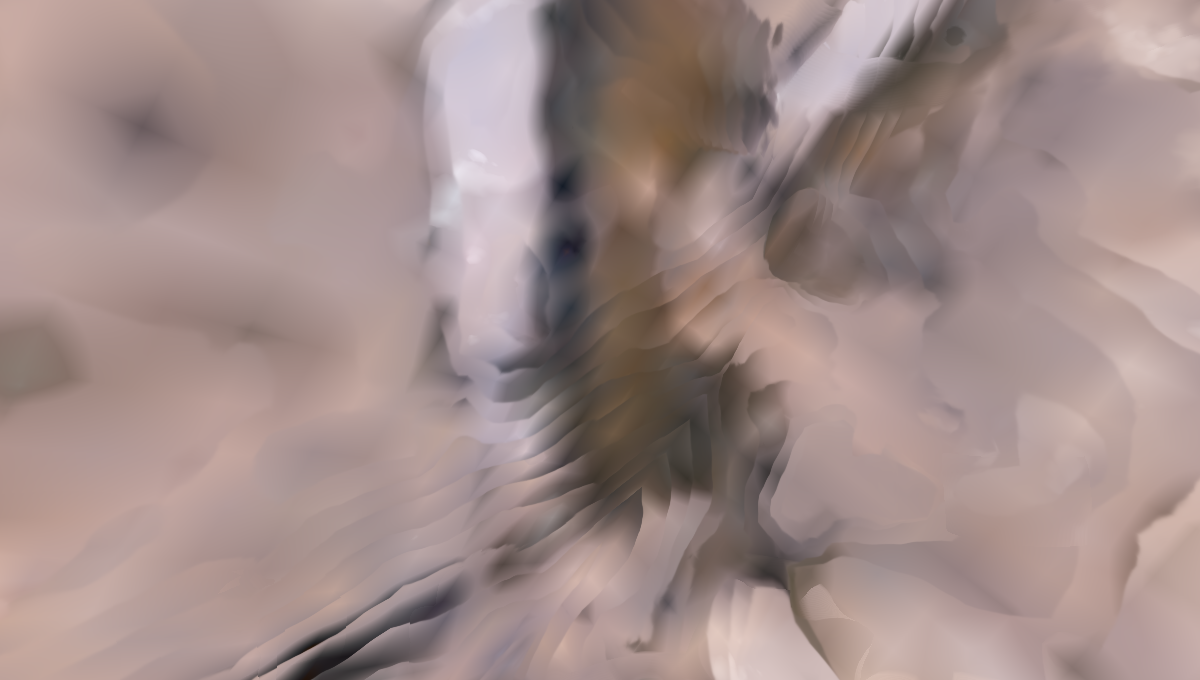}
    \caption{NICE-SLAM \cite{zhu2022nice} with $\tau = 5$ (PSNR: 12.42)}
  \end{subfigure}
  \hfill
  \begin{subfigure}[b]{0.48\linewidth}
    \centering
    \includegraphics[keepaspectratio, scale=0.18]{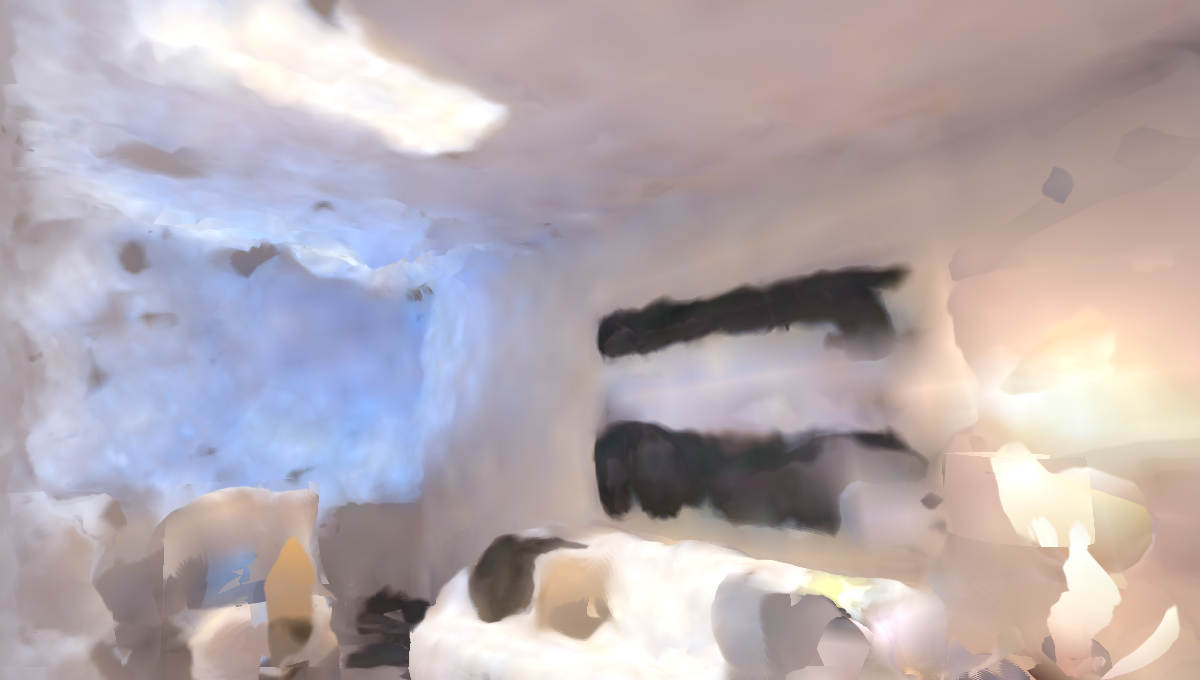}
    \caption{\textbf{EvenNICER-SLAM} with $\tau = 5$ (PSNR: 18.87)}
  \end{subfigure}
  \hfill
  \begin{subfigure}[b]{0.48\linewidth}
    \centering
    \includegraphics[keepaspectratio, scale=0.18]{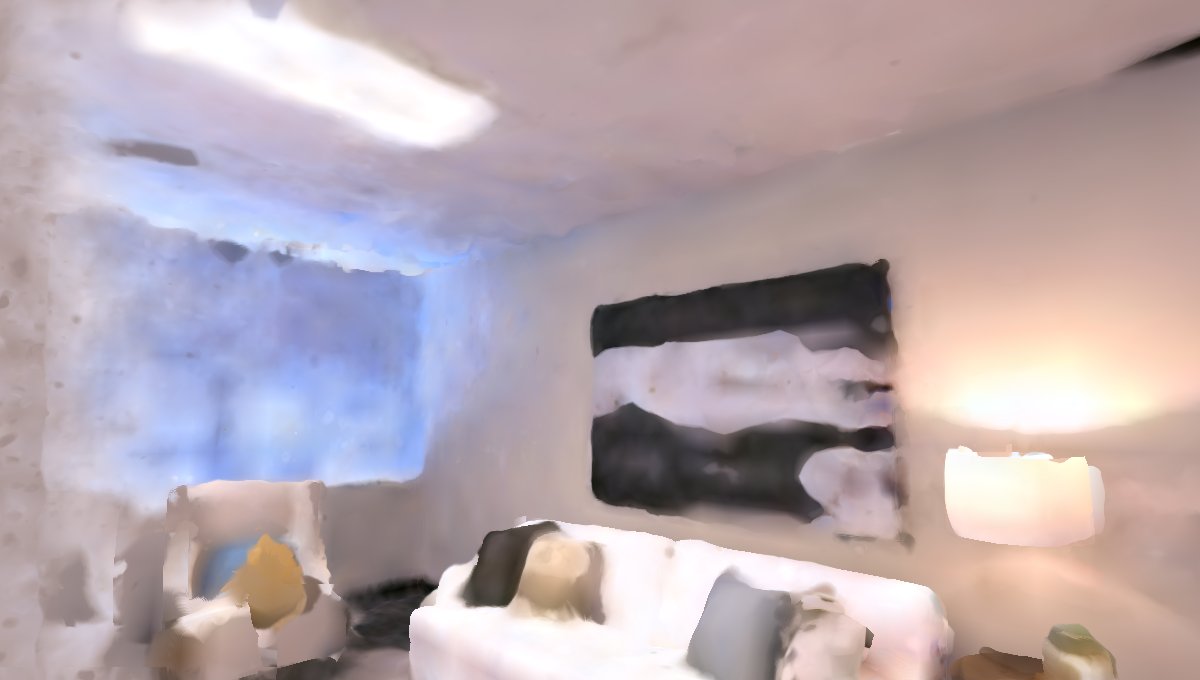}
    \caption{Full-input NICE-SLAM \cite{zhu2022nice}  (PSNR: 21.59)}
  \end{subfigure}
   \caption[Examples: rendered images]{Visual comparisons between a ground truth RGB image (frame \#710 of \texttt{room0} from Replica\cite{straub2019replica}) and the corresponding images rendered by NICE-SLAM \cite{zhu2022nice} and EvenNICER-SLAM both with frame gap $\tau = 5$, and the full-input NICE-SLAM. We also include PSNR values of the rendered images for reference. Note that the (d) full-input NICE-SLAM uses 5 times more RGB-D frames than (b) and (c).}
  \label{fig:render}
\end{figure*}

\section{Experiments and Results}


We evaluate our SLAM framework on a synthetic indoor-scene dataset Replica\cite{straub2019replica}. In this section, we first provide information on our experimental setup for reproducibility. Then, we present both qualitative and quantitative evaluation results and discussions on them.

\subsection{Experimental Setup}
\noindent\textbf{Dataset.} We use Replica \cite{straub2019replica} RGB-D image sequences rendered in 8 different scenes (each 2000 frames long). The sequences are directly downloaded from the NICE-SLAM \cite{zhu2022nice} repository. Event images between adjacent frames are generated by ESIM \cite{rebecq2018esim, Gehrig_2020_CVPR} prior to the experiments.


\noindent\textbf{ESIM.} We use the event simulator ESIM\cite{rebecq2018esim, Gehrig_2020_CVPR} to generate ground truth event images. We fix the parameters of ESIM for all ground truth event images. We use positive and negative contrast threshold of $C_{+} = C_{-} = 0.1$ and a refractory period [s] (the minimum waiting period before a pixel can trigger a new event) of $t_{ref} = 10^{-4}$. We set $\epsilon = 10^{-3}$ in Eq.\eqref{eq:1}.

\noindent\textbf{Event-Net.} We use an $L_{2}$-loss between ground truth and predicted event images to pretrain Event-Net. We do not downscale the images during the training phase. The pretraining process is limited to 1 epoch on the Replica dataset\cite{straub2019replica}. 

\noindent\textbf{Baselines.} We use NICE-SLAM \cite{zhu2022nice} with RGB-D input with frame gaps as baselines. A frame gap 5 (i.e. $\tau = 5$) means four frames out of every five from the original RGB-D stream are excluded. This results in an increase in inter-frame camera translations and rotations, simulating a faster camera motion scenario. 

\noindent\textbf{Metrics.} We employ the same metrics as in NICE-SLAM \cite{zhu2022nice} for evaluation. The \textit{ATE RMSE} [cm] is used to evaluate camera tracking error. For the evaluation of mapping quality, we consider \textit{L1 depth loss} [cm], \textit{accuracy} [cm], \textit{completion} [cm], and \textit{completion ratio} [$<5$cm \%].

\noindent\textbf{Implementation Details.} All experiments are run on either one NVIDIA GeForce GTX TITAN X or one NVIDIA TITAN Xp GPU. For EvenNICER-SLAM, event images are made always available except for the first frame, while RGB-D images are only fed once every $\tau$ frames. The tracker thread uses both RGB-D and event images when the frame ID is a multiple of $\tau$ and only event images otherwise. The mapper threads are dispatched once every $\tau$ frames, and only use RGB-D information for mapping. The scale factor is 0.15 ($1200\times680$ Replica\cite{straub2019replica} images are downscaled to a size of $180\times102$). For Gaussian filtering on event images, we use a $9\times9$ Gaussian kernel with a standard deviation $\sigma = 1.7$ in both $x$ and $y$ directions. Empirically we set $\lambda_{event} = 0.025$.

\begin{table*}[t]
\begin{center}
\begin{tabular}{ccccccccc}
\hline
Scenes                      & room0   & room1   & room2 & office0 & office1 & office2 & office3 & office4 \\\hline
NICE-SLAM\cite{zhu2022nice} & F(701)  & F(380)  & 262.6 & 135     & F(350)  & F(1754) & F(658)  & F(465)  \\
\textbf{EvenNICER-SLAM}     & 28.61   & F(1588) & 6.43  & 37.61   & 47.06   & 11.40   & 13.33   & F(1257) \\
\hline
\end{tabular}
\end{center}
\caption[Quantitative results: camera tracking]{Quantitative evaluation results of camera tracking on Replica\cite{straub2019replica}. The metric adopted here is \textit{ATE RMSE} [cm] ($\downarrow$). An ``F" indicates that the SLAM lost track of the camera and failed to complete the entire sequence. The value in parentheses following the ``F" represents the number of frames out of 2000 that the SLAM system was able to process before crashing.}
\label{table:tracking}
\end{table*}

\begin{table*}[t]
\begin{center}
\begin{tabular}{cccccccccc}
\hline
\multicolumn{2}{c}{Scenes}                                                                  & room0 & room1 & room2  & office0 & office1 & office2 & office3 & office4 \\ \hline
\multirow{4}{*}{NICE-SLAM}      & Depth L1 $\downarrow$    & F     & F     & 86.79  & 78.74   & F       & F       & F       & F       \\
& Accuracy $\downarrow$    & F     & F     & 149.92 & 107.19  & F       & F       & F       & F       \\
& Completion $\downarrow$  & F     & F     & 22.90  & 29.78   & F       & F       & F       & F       \\
& Comp. Ratio $\uparrow$ & F     & F     & 18.35  & 17.56   & F       & F       & F       & F       \\ \hline
\multirow{4}{*}{\textbf{EvenNICER-SLAM}} & Depth L1 $\downarrow$    & 23.23 & F     & 2.17   & 41.12   & 29.44   & 10.19   & 11.27   & F       \\
& Accuracy $\downarrow$    & 18.09 & F     & 2.31   & 31.08   & 22.79   & 4.55    & 5.78    & F       \\
& Completion $\downarrow$  & 13.14 & F     & 2.55   & 19.18   & 17.74   & 3.97    & 5.61    & F       \\
& Comp. Ratio $\uparrow$ & 41.26 & F     & 90.65  & 22.30   & 38.32   & 76.71   & 62.07   & F       \\ \hline
\end{tabular}
\end{center}
\caption[Quantitative results: mapping]{Quantitative evaluation results of mapping on Replica\cite{straub2019replica}. The metrics are \textit{L1 depth loss} [cm], \textit{accuracy} [cm], \textit{completion} [cm], \textit{completion ratio} [\%]. Again, an ``F" stands for a failure of the SLAM system.}
\label{table:mapping}
\end{table*}

\subsection{Results}
\noindent\textbf{Qualitative Results.}
We first give a visual comparison of predicted camera trajectories from NICE-SLAM \cite{zhu2022nice} and EvenNICER-SLAM in Figure \ref{fig:traj}. We can see that while NICE-SLAM with frame gap $\tau = 5$ constantly experiences a deviation from the ground truth camera poses, EvenNICER-SLAM predicts a similar camera pose most of the time and shows a stronger capability of fixing minor tracking errors. In Figure \ref{fig:render}, we present examples of rendered RGB images when frame gap $\tau = 5$, providing a qualitative evaluation of scene mapping. Additionally, we calculate the PSNR measurements of these rendered images with regard to the ground truth. We can see that while the full-input NICE-SLAM gives the best reconstruction result, our EvenNICER-SLAM also produces plausible mapping and achieves a similar PSNR value. Note that EvenNICER-SLAM achieves this result under the constraint of only using a fifth of the original RGB-D input frames. Also, intuitively we can tell that EvenNICER-SLAM keeps track of the camera well, while the NICE-SLAM with reduced input loses track of the camera completely and the mapping is severely distorted.

\begin{figure}[t]
\centering
\includegraphics[scale=0.16]{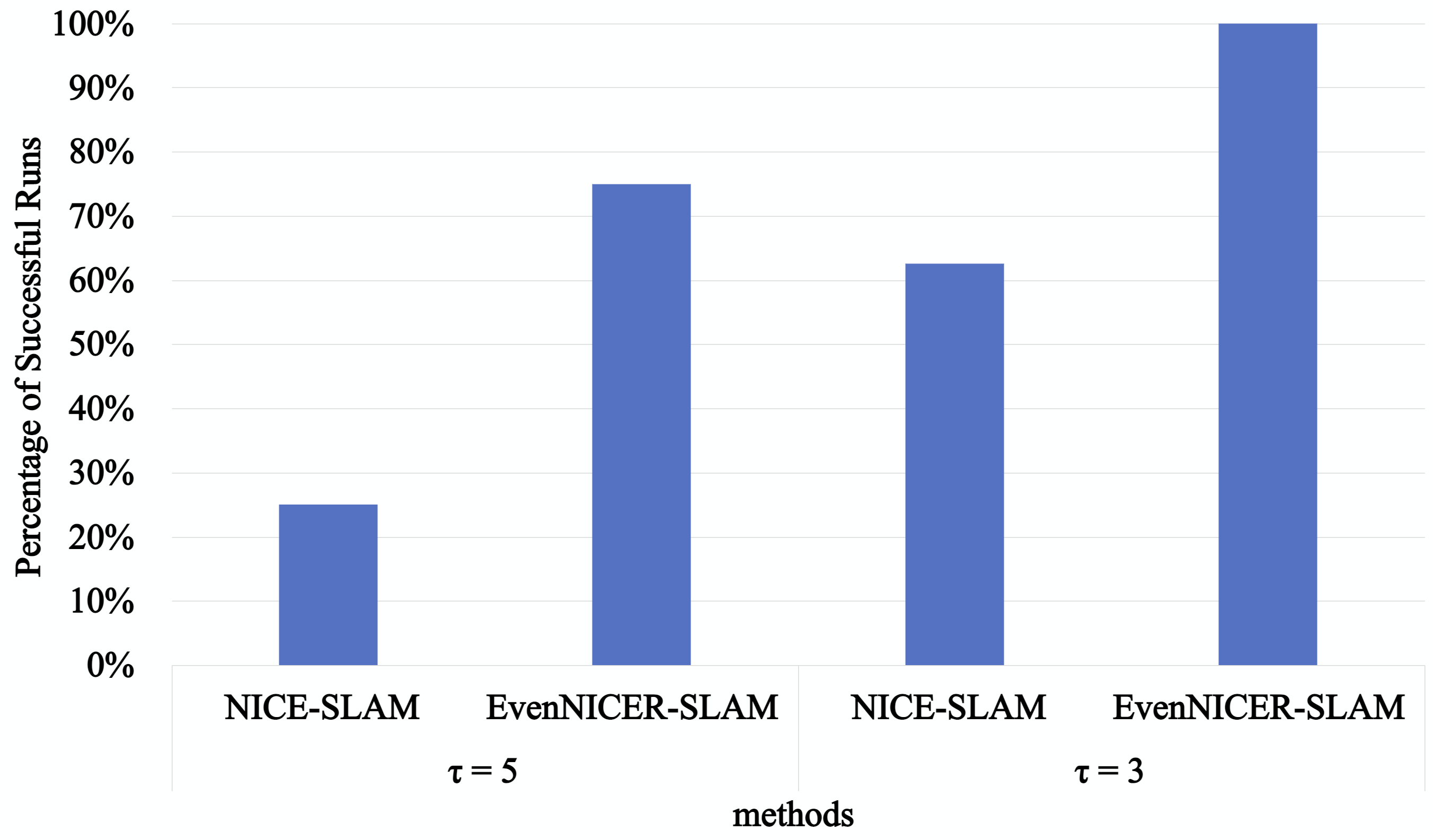}
\caption[Success Rates]{A comparison of success rates of NICE-SLAM\cite{zhu2022nice} and EvenNICER-SLAM under different frame gaps. With the same RGB-D input frequency, EvenNICER-SLAM significantly reduces the chance of crashing.}
  \label{fig:chart}
\end{figure}

\noindent\textbf{Quantitative Results.}
We present quantitative results on camera tracking and mapping in Table \ref{table:tracking} and Table \ref{table:mapping} respectively. These tables provide detailed numerical evaluations of the performance of each method in terms of tracking accuracy and mapping quality. The comparisons of our EvenNICER-SLAM and NICE-SLAM in Table \ref{table:tracking} shows that the presence of event supervision largely reduces camera tracking error. We observe a similar tendency in terms of mapping quality from Table \ref{table:mapping}. Since we keep the mapping process consistent, this observation indicates a positive correlation between camera tracking accuracy and mapping quality. In Figure \ref{fig:chart}, we plot an additional histogram illustrating the success rates of the SLAM systems under different settings based on our experimental results. The vertical axis represents the percentage of successful runs that complete a whole 2000-frame Replica\cite{straub2019replica} sequence. We can see that EvenNICER-SLAM significantly lowers the chance of losing track of the camera. With three frame gaps ($\tau = 3$), we observed no failure of our method.

\noindent\textbf{Interpretation.}
From the qualitative and quantitative results, we can first conclude that event information can positively contribute to camera tracking. Additionally, better tracking leads to better mapping. High-frequency event input can also indirectly contribute to an improved mapping quality with lower-frequency RGB-D input. It is difficult for EvenNICER-SLAM with reduced RGB-D frames to match the performance of the original full-input NICE-SLAM \cite{zhu2022nice} as an event image essentially provides less information than a set of RGB-D images. Nevertheless, our experimental results suggest the possibility that EvenNICER-SLAM can outperform the original NICE-SLAM with the same RGB-D input frequency and a higher-frequency event input.

\begin{figure}[t]
\centering
\includegraphics[scale=0.57]{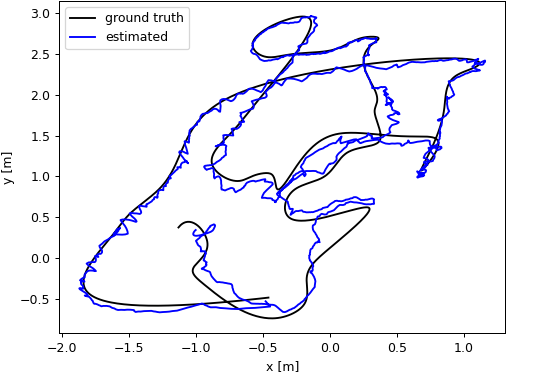}
\caption[Limitations]{A typical trajectory predicted by EvenNICER-SLAM (with $\tau = 5$ in \texttt{office2} from Replica\cite{straub2019replica}). Despite not showing large deviations, EvenNICER-SLAM tends to predict zigzag trajectories around ground truth camera poses and rarely converges to a smooth trajectory.}
  \label{fig:limitation}
\end{figure}

\section{Discussion}
In this section, we first discuss the limitations of our methods and try to propose potential approaches to improving the results in the current problem setting. Then, we further explore the possibility of EvenNICER-SLAM and its potential applications.


\subsection{Limitations}
In Figure \ref{fig:limitation}, we show a typical example of trajectories predicted by EvenNICER-SLAM. We can see that EvenNICER-SLAM tends to generate zigzag trajectories around the ground truth camera poses and struggles to converge to a smooth trajectory. While this does not directly result in significant camera tracking errors under the \textit{ATE RMSE} metric adopted in our experiments, it can adversely impact mapping quality and consequently degrade tracking performance. According to our observation, a zigzag predicted trajectory may cause wavy artifacts in rendered images due to the interdependence of tracking and mapping. These mapping artifacts would further lead to artifacts in event predictions and potentially mislead the camera pose optimization process.

Despite the Gaussian blurring solution proposed in Sec. \ref{sec:optim}, the optimization based on event loss remains relatively unstable compared to the photometric loss utilized in NICE-SLAM\cite{zhu2022nice}. While the event loss based on downscaled event images enables EvenNICER-SLAM to correct larger tracking errors at an affordable computational cost, it may also cause random moves in the parameter space when errors are small. We anticipate that a coarse-to-fine optimization paradigm would help further alleviate the alignment issue. Namely, we can employ a hierarchical optimization pipeline using larger Gaussian kernels initially and reducing the kernel size subsequently.

\subsection{Future Work}
Furthermore, the use of event loss in the mapper thread still remains unexplored. If EvenNICER-SLAM could gain meaningful information about scene geometry from event input, it would even be possible to avoid the depth map input. Such extensions would greatly expand the range of applicable synthetic datasets and reduce the cost of data collection. One possible approach is first estimating dense optical flow from events, as proposed in \cite{gehrig2021raft}. The optical flow estimation can then be integrated into the SLAM system, as demonstrated in the case of DROID-SLAM \cite{teed2021droid}.

Another future work  for EvenNICER-SLAM is to use it in the real domain. The low-latency and no-motion-blur nature of event cameras could help SLAM systems work well even with fast camera motions. A potential issue is that real event data would be much noisier than ESIM \cite{rebecq2018esim, Gehrig_2020_CVPR} simulations. To this end, adversarial training of Event-Net might be helpful in closing the domain gap between synthetic data and real data. If EvenNICER-SLAM is successfully adapted to the real domain, it could help, for example, autonomous drones and self-driving vehicles perceive their surrounding environments and plan their moves better.

\section{Conclusion}
In this paper, we proposed EvenNICER-SLAM, a novel event-based neural implicit encoding SLAM system. We have developed the pipeline by integrating event image input into the existing method NICE-SLAM \cite{zhu2022nice}, enabling camera tracking with reduced RGB-D input. Our quantitative evaluation of EvenNICER-SLAM and NICE-SLAM in both tracking and mapping has revealed that higher-frequency feed of event images greatly enhances tracking and mapping quality, particularly when RGB-D input is temporally sparse. This has been made possible thanks to the proposed Event-Net module. While designing the Event-Net module, both differentiability and transferability aspects were considered. Although not exploited in the current stage, the transferability aspect of the Event-Net will potentially facilitate knowledge transfer across domains. We hope this will make learning from synthetic data and transferring that knowledge via domain adaptation possible.
{\small
\bibliographystyle{ieee_fullname}
\bibliography{egbib}

\begin{thebibliography}{10}\itemsep=-1pt

\bibitem{bloesch2018codeslam}
Michael Bloesch, Jan Czarnowski, Ronald Clark, Stefan Leutenegger, and Andrew~J
  Davison.
\newblock Codeslam—learning a compact, optimisable representation for dense
  visual slam.
\newblock In {\em Proceedings of the IEEE conference on computer vision and
  pattern recognition}, pages 2560--2568, 2018.

\bibitem{gallego2020event}
Guillermo Gallego, Tobi Delbr{\"u}ck, Garrick Orchard, Chiara Bartolozzi, Brian
  Taba, Andrea Censi, Stefan Leutenegger, Andrew~J Davison, J{\"o}rg Conradt,
  Kostas Daniilidis, et~al.
\newblock Event-based vision: A survey.
\newblock {\em IEEE transactions on pattern analysis and machine intelligence},
  44(1):154--180, 2020.

\bibitem{Gehrig_2020_CVPR}
Daniel Gehrig, Mathias Gehrig, Javier Hidalgo-Carri\'o, and Davide Scaramuzza.
\newblock Video to events: Recycling video datasets for event cameras.
\newblock In {\em {IEEE} Conf. Comput. Vis. Pattern Recog. (CVPR)}, June 2020.

\bibitem{gehrig2021raft}
Mathias Gehrig, Mario Millh{\"a}usler, Daniel Gehrig, and Davide Scaramuzza.
\newblock E-raft: Dense optical flow from event cameras.
\newblock In {\em 2021 International Conference on 3D Vision (3DV)}, pages
  197--206. IEEE, 2021.

\bibitem{guan2022deepmix}
Yongjie Guan, Xueyu Hou, Nan Wu, Bo Han, and Tao Han.
\newblock Deepmix: mobility-aware, lightweight, and hybrid 3d object detection
  for headsets.
\newblock In {\em Proceedings of the 20th Annual International Conference on
  Mobile Systems, Applications and Services}, pages 28--41, 2022.

\bibitem{hanover2023autonomous}
Drew Hanover, Antonio Loquercio, Leonard Bauersfeld, Angel Romero, Robert
  Penicka, Yunlong Song, Giovanni Cioffi, Elia Kaufmann, and Davide Scaramuzza.
\newblock Autonomous drone racing: A survey.
\newblock {\em arXiv e-prints}, pages arXiv--2301, 2023.

\bibitem{kim2016real}
Hanme Kim, Stefan Leutenegger, and Andrew~J Davison.
\newblock Real-time 3d reconstruction and 6-dof tracking with an event camera.
\newblock In {\em European conference on computer vision}, pages 349--364.
  Springer, 2016.

\bibitem{klein2009parallel}
Georg Klein and David Murray.
\newblock Parallel tracking and mapping on a camera phone.
\newblock In {\em 2009 8th IEEE International Symposium on Mixed and Augmented
  Reality}, pages 83--86. IEEE, 2009.

\bibitem{lin2021barf}
Chen-Hsuan Lin, Wei-Chiu Ma, Antonio Torralba, and Simon Lucey.
\newblock Barf: Bundle-adjusting neural radiance fields.
\newblock In {\em Proceedings of the IEEE/CVF International Conference on
  Computer Vision}, pages 5741--5751, 2021.

\bibitem{mildenhall2021nerf}
Ben Mildenhall, Pratul~P Srinivasan, Matthew Tancik, Jonathan~T Barron, Ravi
  Ramamoorthi, and Ren Ng.
\newblock Nerf: Representing scenes as neural radiance fields for view
  synthesis.
\newblock {\em Communications of the ACM}, 65(1):99--106, 2021.

\bibitem{newcombe2011dtam}
Richard~A Newcombe, Steven~J Lovegrove, and Andrew~J Davison.
\newblock Dtam: Dense tracking and mapping in real-time.
\newblock In {\em 2011 international conference on computer vision}, pages
  2320--2327. IEEE, 2011.

\bibitem{rebecq2018esim}
Henri Rebecq, Daniel Gehrig, and Davide Scaramuzza.
\newblock Esim: an open event camera simulator.
\newblock In {\em Conference on robot learning}, pages 969--982. PMLR, 2018.

\bibitem{rebecq2016evo}
Henri Rebecq, Timo Horstsch{\"a}fer, Guillermo Gallego, and Davide Scaramuzza.
\newblock Evo: A geometric approach to event-based 6-dof parallel tracking and
  mapping in real time.
\newblock {\em IEEE Robotics and Automation Letters}, 2(2):593--600, 2016.

\bibitem{ronneberger2015u}
Olaf Ronneberger, Philipp Fischer, and Thomas Brox.
\newblock U-net: Convolutional networks for biomedical image segmentation.
\newblock In {\em International Conference on Medical image computing and
  computer-assisted intervention}, pages 234--241. Springer, 2015.

\bibitem{rosinol2022nerf}
Antoni Rosinol, John~J Leonard, and Luca Carlone.
\newblock Nerf-slam: Real-time dense monocular slam with neural radiance
  fields.
\newblock {\em arXiv preprint arXiv:2210.13641}, 2022.

\bibitem{schops2019bad}
Thomas Schops, Torsten Sattler, and Marc Pollefeys.
\newblock Bad slam: Bundle adjusted direct rgb-d slam.
\newblock In {\em Proceedings of the IEEE/CVF Conference on Computer Vision and
  Pattern Recognition}, pages 134--144, 2019.

\bibitem{straub2019replica}
Julian Straub, Thomas Whelan, Lingni Ma, Yufan Chen, Erik Wijmans, Simon Green,
  Jakob~J Engel, Raul Mur-Artal, Carl Ren, Shobhit Verma, et~al.
\newblock The replica dataset: A digital replica of indoor spaces.
\newblock {\em arXiv preprint arXiv:1906.05797}, 2019.

\bibitem{sucar2021imap}
Edgar Sucar, Shikun Liu, Joseph Ortiz, and Andrew~J Davison.
\newblock imap: Implicit mapping and positioning in real-time.
\newblock In {\em Proceedings of the IEEE/CVF International Conference on
  Computer Vision}, pages 6229--6238, 2021.

\bibitem{sucar2020nodeslam}
Edgar Sucar, Kentaro Wada, and Andrew Davison.
\newblock Nodeslam: Neural object descriptors for multi-view shape
  reconstruction.
\newblock In {\em 2020 International Conference on 3D Vision (3DV)}, pages
  949--958. IEEE, 2020.

\bibitem{teed2021droid}
Zachary Teed and Jia Deng.
\newblock Droid-slam: Deep visual slam for monocular, stereo, and rgb-d
  cameras.
\newblock {\em Advances in Neural Information Processing Systems},
  34:16558--16569, 2021.

\bibitem{ungureanu2020hololens}
Dorin Ungureanu, Federica Bogo, Silvano Galliani, Pooja Sama, Xin Duan, Casey
  Meekhof, Jan St{\"u}hmer, Thomas~J Cashman, Bugra Tekin, Johannes~L
  Sch{\"o}nberger, et~al.
\newblock Hololens 2 research mode as a tool for computer vision research.
\newblock {\em arXiv preprint arXiv:2008.11239}, 2020.

\bibitem{vidal2018ultimate}
Antoni~Rosinol Vidal, Henri Rebecq, Timo Horstschaefer, and Davide Scaramuzza.
\newblock Ultimate slam? combining events, images, and imu for robust visual
  slam in hdr and high-speed scenarios.
\newblock {\em IEEE Robotics and Automation Letters}, 3(2):994--1001, 2018.

\bibitem{yen2021inerf}
Lin Yen-Chen, Pete Florence, Jonathan~T Barron, Alberto Rodriguez, Phillip
  Isola, and Tsung-Yi Lin.
\newblock inerf: Inverting neural radiance fields for pose estimation.
\newblock In {\em 2021 IEEE/RSJ International Conference on Intelligent Robots
  and Systems (IROS)}, pages 1323--1330. IEEE, 2021.

\bibitem{zhi2019scenecode}
Shuaifeng Zhi, Michael Bloesch, Stefan Leutenegger, and Andrew~J Davison.
\newblock Scenecode: Monocular dense semantic reconstruction using learned
  encoded scene representations.
\newblock In {\em Proceedings of the IEEE/CVF Conference on Computer Vision and
  Pattern Recognition}, pages 11776--11785, 2019.

\bibitem{zhu2022nice}
Zihan Zhu, Songyou Peng, Viktor Larsson, Weiwei Xu, Hujun Bao, Zhaopeng Cui,
  Martin~R Oswald, and Marc Pollefeys.
\newblock Nice-slam: Neural implicit scalable encoding for slam.
\newblock In {\em Proceedings of the IEEE/CVF Conference on Computer Vision and
  Pattern Recognition}, pages 12786--12796, 2022.

\end{thebibliography}
}

\end{document}